\documentclass{article}

\usepackage{microtype}
\usepackage{graphicx}
\usepackage{subcaption}
\usepackage{booktabs} 

\usepackage{hyperref}

\usepackage[accepted]{icml2026}

\usepackage{amsmath}
\usepackage{amssymb}
\usepackage{mathtools}
\usepackage{amsthm}

 \usepackage{multirow}

\usepackage[capitalize,noabbrev]{cleveref}

\theoremstyle{plain}

\theoremstyle{definition}

\theoremstyle{remark}

\usepackage[textsize=tiny]{todonotes}

\icmltitlerunning{DELBERT-2: Pretrained Fingerprint Language Models for DEL Protein Binder Prediction}

\begin{document}

\twocolumn[
  \icmltitle{DELBERT-2: Pretrained Fingerprint Language Models for DEL Protein Binder Prediction}



  \icmlsetsymbol{equal}{*}

  \begin{icmlauthorlist}
    \icmlauthor{Bing Hu}{cs}
    \icmlauthor{Sun Sun}{nrc}
    \icmlauthor{Shaik Salman Basha}{cs}
    \icmlauthor{Anita Layton}{am}
    \icmlauthor{Helen Chen}{ph}
  \end{icmlauthorlist}

  \icmlaffiliation{cs}{Computer Science, University of Waterloo, Canada}
  \icmlaffiliation{nrc}{National Research Council of Canada, Canada}
  \icmlaffiliation{am}{Applied Mathematics, University of Waterloo, Canada}
  \icmlaffiliation{ph}{School of Public Health, University of Waterloo, Canada}

  \icmlcorrespondingauthor{Bing Hu}{b25hu@uwaterloo.ca}

  \icmlkeywords{Molecular Foundation Models, Chemical Fingerprints, Protein Binder Prediction}

  \vskip 0.3in
]



\printAffiliationsAndNotice{}  

\begin{abstract}
We present DELBERT-2, a pretrained fingerprint language model that converts sparse molecular fingerprints into unified token sequences using a ModernBERT encoder trained with masked language modelling on 2.5M molecules from the AIRCHECK DEL corpus.
We evaluate across six targets (WDR91, WDR12, SETDB1, PLCZ1, LRRK2, DCAF7) under three out-of-distribution (OOD) protocols: hierarchical cluster splits probe chemical novelty, library splits test cross-library transfer, and building-block splits assess compositional generalization.
DELBERT-2 consistently improves PR-AUC and NDCG@1000 relative to LightGBM ensemble baselines and transformers trained from scratch, with the largest gains in stringent OOD regimes.
DELBERT-2 achieves 13.28±3.72 enrichment factor at top-100 vs. 12.36±3.75 for no-pretraining (p=0.040), representing 7.4\% improvement and 13× enrichment over random selection.
These results demonstrate that fingerprint-centric self-supervised learning effectively improves hit prioritization under distribution shift, enabling practical DEL virtual screening. Code is available at \hyperlink{https://github.com/GenerativeDrugDiscovery/DELBERT-2}{DELBERT-2}.
\end{abstract}
\section{Introduction}

Hit identification remains a major bottleneck in early-stage drug discovery: the goal is to find small molecules that bind a desired protein target with sufficient quality to justify downstream optimization \cite{wellnitz2024enabling}.
For many years, high-throughput screening using functional assays has been the predominant approach for discovering small-molecule hits \cite{blay2020high}. 
One of the more cost-effective technologies to emerge is the DNA-encoded library (DEL), in which each compound is tagged with a unique DNA barcode that encodes its synthetic history and composition \cite{brenner1992encoded}.  
DELs facilitate the rapid screening of billions of compounds at relatively low cost \cite{gironda2021dna}. 


At the same time, DEL screening has produced increasingly large public datasets, including those released through the Artificial Intelligence Ready Chemical Knowledge (AIRCHECK) database \cite{reza2026aircheck}.
These resources enable machine learning models for DEL virtual screening, prioritizing candidates for experimental follow-up without costly resynthesis.
However, data scarcity remains a fundamental challenge \cite{mswahili2024transformer, wellnitz2024enabling}, and current DEL-ML methods often fail on compounds that differ substantially from the training distribution—including molecules from unseen libraries, unseen building blocks, or chemically novel regions of space \cite{quigley2024belka, wellnitz2024enabling}.
Improving out-of-distribution (OOD) generalization is therefore critical for making DEL-ML practically useful.


\begin{figure*}[ht]
    \centering
    \includegraphics[width=1\linewidth]{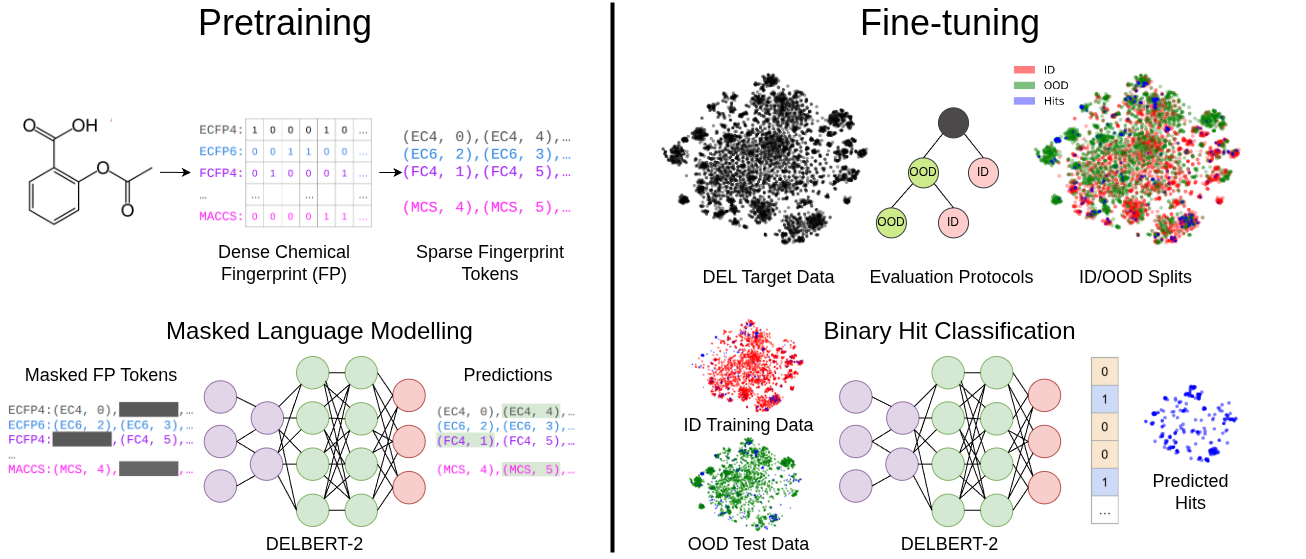}
    \caption{DELBERT-2 pre-training and fine-tuning workflow. Multiple sparse fingerprints are tokenized into a unified sequence: each active bit at position \texttt{pos} in fingerprint type \texttt{fp} becomes a discrete token \texttt{fp\_pos\_val}, transforming sparse binary vectors into variable-length token sequences. The model is pre-trained with masked language modeling (MLM) on 2.5M DEL molecules, then fine-tuned with LoRA for target-specific binder prediction.}
    \label{fig:arch}
\end{figure*}

Transformers have emerged as a powerful architecture for learning self-supervised representations, where pretraining followed by task-specific fine-tuning can provide substantial gains on downstream tasks, especially under distribution shift \cite{hendrycks-etal-2020-pretrained}. 
In molecular machine learning, much of this progress has focused on SMILES strings or graph representations. However, in practice, many strong DEL screening pipelines continue to rely on molecular fingerprints because they are efficient, chemically informative, and widely used in both public and industrial workflows \cite{cereto2015molecular, mswahili2024transformer}.
Furthermore, public DEL datasets such as AIRCHECK release only precomputed molecular fingerprints to preserve the confidentiality of proprietary chemical structures, making fingerprint-only approaches essential for broad collaboration.

Recent work explored fingerprint language modelling for DEL hit prediction \cite{delbert}.
Building on this, we introduce DELBERT-2, which extends the framework through \emph{unified pre-training} on the entire AIRCHECK DEL corpus (2.5M molecules vs.\ 500K per-target) and comprehensive OOD evaluation across three complementary protocols (\autoref{fig:arch}).
Across six AIRCHECK targets, DELBERT-2 consistently improves PR-AUC and NDCG@1000 relative to strong LightGBM baselines and transformers trained from scratch, with the clearest gains in more stringent OOD regimes.

  
Our contributions are as follows:  
\begin{itemize}  
    \item We demonstrate that unified pretraining on 2.5M molecules from the entire AIRCHECK DEL corpus enables better cross-target transfer learning compared to no pretraining approaches.
    \item We introduce three complementary OOD evaluation protocols (hierarchy, library, building-block splits) that systematically probe chemical novelty, cross-library transfer, and compositional generalization—key failure modes for practical DEL virtual screening.
    \item We show that self-supervised pretraining over fingerprint language leads to improved DEL binder prediction across six high-quality targets and quantify practical impact through enrichment analysis, achieving 13$\times$ enrichment at top-100.
\end{itemize}  

\section{Background}

\textbf{LMs for Chemistry.} SMILES strings \cite{weininger1988smiles} are textual representations of molecular structures that have enabled the application of NLP methods to accelerate drug discovery in various applications, such as molecular property prediction, drug-target interaction prediction, and hit generation \cite{yu2021review, wang2020predicting, hu2024synthetic, hu2025drug, ahmad2023discovery}.
In particular, transformer-based models trained on SMILES, such as MOLBERT \cite{li2021mol}, ChemBERTa \cite{chithrananda2020chemberta}, and MoLFormer \cite{wu2023molformer}, have made significant advances in the field of computational chemistry \cite{mswahili2024transformer}.
Further advances in data efficiency under limited labeled data settings can improve performance, particularly on complex tasks or rare molecular properties \cite{mswahili2024transformer}. 
Pre-training of transformer-based models builds a deep understanding of relationships, patterns, and features within the data structures, enabling the model to perform well on broad range of in-distribution and out-of-distribution downstream tasks \cite{hendrycks-etal-2020-pretrained}. 

\textbf{Fingerprint Representations.} Molecular fingerprints can be computed from SMILES strings, where each method aims to capture and encode different aspects of a molecule \cite{cereto2015molecular}. 
Virtual screening \cite{rester2008virtuality} methods using molecular fingerprints leverage molecular similarity to known hits and non-hits. 
There are several types of molecular fingerprints, including substructure key-based fingerprints \cite{durant2002reoptimization}, path-based fingerprints \cite{dalke2019chemfp}, and circular topological fingerprints \cite{morgan1965generation}. 
Fingerprint representations are widely used in DEL hit discovery pipelines, with large-scale fingerprint datasets published on the AIRCHECK platform \cite{edwards2025protein}. 
Importantly, the fingerprint-only setting introduces unique challenges: it precludes the use of existing molecular encoders that depend on structural information (including graph neural networks and SMILES-based transformers), and existing fingerprint-based approaches rely on supervised tree-based models that cannot leverage unlabeled data for self-supervised learning.

\textbf{DEL-ML Hit Generation.} Virtual screening models, trained from DEL libraries, forming a DEL-ML pipeline \cite{mccloskey2020machine, iqbal2024del+, han2024highly}, can be used to screen even larger commercial libraries of small molecules, enabling the selection, purchase, and testing of hit compounds \cite{quigley2024belka}.
The DEL-ML pipeline avoids the need for resynthesis of original library hits and enables filtering of hit compounds based on chemical properties prior to selection and purchase \cite{wellnitz2024enabling}.
With fingerprint DEL data from AIRCHECK, \citet{wellnitz2024enabling} show tree-based and boosting ensemble methods find success in a virtual screening task predicting hits for target WDR91. 
Leveraging fingerprints, alongside with SMILES, from BELKA \cite{quigley2024belka}, \citet{shlepov2024belka} show that small transformer architectures can be fine-tuned to predict hits for a set of 3 targets.
A key limitations remains in OOD prediction, where current models fail to effectively generalize to chemicals distinct from the training set \cite{leash-BELKA, wellnitz2024enabling}.

\begin{figure*}[ht]
    \centering
    \includegraphics[width=1\linewidth]{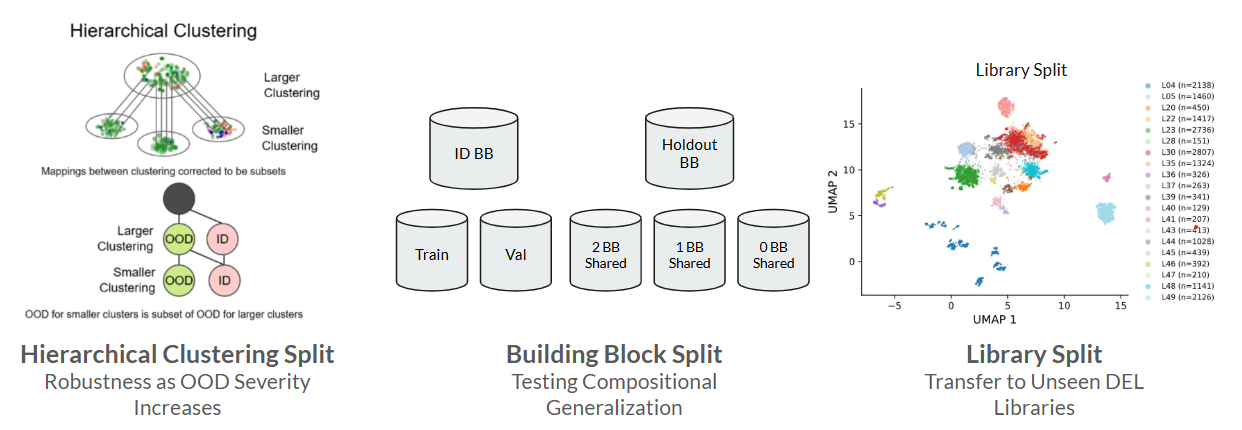}
    \caption{Out-of-distribution (OOD) evaluation protocols for DEL virtual screening.
    \textbf{Left: Hierarchical clustering split} constructs nested ID/OOD partitions at multiple Tanimoto similarity \citep{tanimoto1958elementary} cutoffs (0.65/0.75/0.85); higher cutoffs create tighter clusters, making OOD test molecules more chemically distant from training.
    \textbf{Middle: Building-block split} holds out 40\% of building blocks (BBs) and stratifies test molecules by BB overlap with training (\texttt{0bb}/\texttt{1bb}/\texttt{2bb}), directly probing compositional generalization.
    \textbf{Right: Library split} holds out entire DEL libraries, testing transfer to unseen synthesis designs (UMAP colored by library; $n$ = number of molecules per library).}
    \label{fig:val_protocols}
\end{figure*}

\section{Methods}

We propose DELBERT-2, a foundation model for DEL virtual screening that learns representations from molecular fingerprints. 
DELBERT-2 is pretrained with self-supervised objectives on large collections of unlabeled molecules and then fine-tuned for downstream DEL hit prediction. 
The central idea is to convert sparse fingerprint vectors into a compact token sequence (a \emph{fingerprint language}) suitable for transformer-based masked language modelling.
See Appendix~\ref{appendix:delbert_comparison} for a detailed comparison of DELBERT vs DELBERT-2 methodology.

We pretrain on approximately 2.5M unlabeled molecules drawn from public DEL collections in the AIRCHECK database \cite{edwards2025protein}. 
Unlike per-target pretraining approaches, this unified approach enables the model to learn generalizable molecular patterns across diverse protein targets and library designs, facilitating better cross-target transfer learning.

\subsection{Chemical Fingerprints}

\paragraph{Fingerprint Tokenization.}
We use molecular fingerprints rather than SMILES because AIRCHECK releases only pre-computed fingerprints to preserve confidentiality of proprietary structures, precluding SMILES-based encoders \cite{cereto2015molecular}.
Given pre-computed fingerprints (ECFP4/6, FCFP4/6, AtomPair, TopTor, and MACCS), we represent each active bit as a discrete token $\texttt{fp\_pos\_val}$, where $\texttt{fp}$ is the fingerprint type, $\texttt{pos}$ is the active bit index, and $\texttt{val}$ is its value.
This transforms high-dimensional sparse vectors into variable-length token sequences suitable for transformer-based masked language modelling.

\textbf{ECFP4} and \textbf{ECFP6}, are extended-connectivity fingerprint with diameter 4 and 6, respectively. They are circular Morgan fingerprints that encode the local circular environment around each atom up to some number of bonds away \cite{morgan1965generation}. In ECFP, atoms are labeled with detailed atomic invariants, such as element, valence, and connectivity, before being hashed into fingerprint bits. 
\textbf{FCFP4} and \textbf{FCFP6}, Functional-Class Fingerprint, are similar circular fingerprints to ECFP, but instead label atoms using functional classes, such as hydrogen-bond donor/acceptor, aromatics, and ionization, before being hashed to fingerprint bits. 
In general, ECFP is better suited for similarity searching and clustering, while FCFP is more appropriate for pharmacophoric applications.

\textbf{AtomPair} fingerprints are topological fingerprints that enumerate all pairs of atoms in a molecule, encoding the atom identifier at each end, and the topological distance, the number of bonds in the shortest path \cite{carhart1985atom}. 
These atom-pair descriptions are then hashed into a fixed-length representation, capturing the global 2D shape of the molecule in terms of atom types and the distance between atoms. 
\textbf{TopTor} is another type of topological fingerprint, aimed at measuring the topological torsion in a molecule \cite{nilakantan1987topological}.
Each torsion is defined as the atom and bond type of four consecutively bonded atoms, before being hashed to fingerprint bits, thereby emphasizing the contiguous fragments in the molecule.

\textbf{MACCS} fingerprints use fixed dictionary-based structural keys to encode molecules based on the presence or absence of predefined substructures or properties \cite{durant2002reoptimization}. MACCS fingerprints are interpretable and are widely used in molecular similarity, clustering, and Quantitative Structure–Activity Relationship (QSAR) modelling \cite{cherkasov2014qsar}.

\subsection{Architecture and Training}

Like DELBERT \cite{delbert}, DELBERT-2 uses a BERT-style masked language modeling encoder; key improvements are: (i) \textit{Unified pre-training:} 2.5M molecules across all AIRCHECK targets vs.\ 500K per-target; (ii) \textit{Expanded fingerprints:} 8 types vs.\ 4; (iii) \textit{Larger backbone:} ModernBERT \cite{warner2024smarterbetterfasterlonger} with 100M parameters (hidden 1024, 12 layers) vs.\ 70M (hidden 640, 16 layers); (iv) \textit{OOD evaluation:} three protocols across 6 targets vs.\ library splits on 4.


\paragraph{Model Architecture.}
DELBERT-2 is a 12-layer ModernBERT encoder \cite{warner2024smarterbetterfasterlonger} with hidden size 1024 (\,$\sim$100M parameters\,), followed by pooling and a binary classification head for fine-tuning.
ModernBERT incorporates Flash Attention, rotary position embeddings (RoPE), and alternating local-global attention layers for efficient processing of long molecular sequences.

\paragraph{Self-supervised Pre training.}
Although AIRCHECK molecules carry binding activity labels, pre-training is entirely self-supervised on fingerprint structures. Fingerprint structures are used for masked language modelling (MLM, 15\% masking, 20 epochs) on $\approx$2.5M molecules \cite{edwards2025protein}.
This unified strategy trains a single model on all available DEL data, enabling transferable representations across diverse targets and library designs.

\paragraph{Fine-tuning.} 
For DEL hit prediction, we fine-tune using LoRA \cite{Hu2021LoRALA} (rank 8, $\alpha{=}16$, applied to \texttt{Wqkv}), which freezes pre-trained weights and injects trainable low-rank matrices for parameter-efficient adaptation.
We optimize with focal loss ($\gamma{=}2$) \cite{lin2017focal}, which down-weights easy negatives to address the 5--11\% positive class rate. Full hyperparameters are in Appendix~\ref{appendix:hps}.

\paragraph{Ensemble LGBM baselines.}
We train LightGBM (LGBM) baselines on the same fingerprint features and train/validation/test splits used for DELBERT-2 to ensure fair comparison. 
We use an ensemble of $K{=}5$ LGBM models per run. Specifically, we apply stratified $K$-fold splitting to the training set and train one LGBM model on each fold's training partition, using the corresponding held-out partition for early stopping; final predictions are obtained by averaging predicted probabilities across ensemble members.

\begin{figure*}[ht]
    \centering
    \begin{minipage}{0.58\textwidth}
        \raggedleft \includegraphics[width=0.95\linewidth]{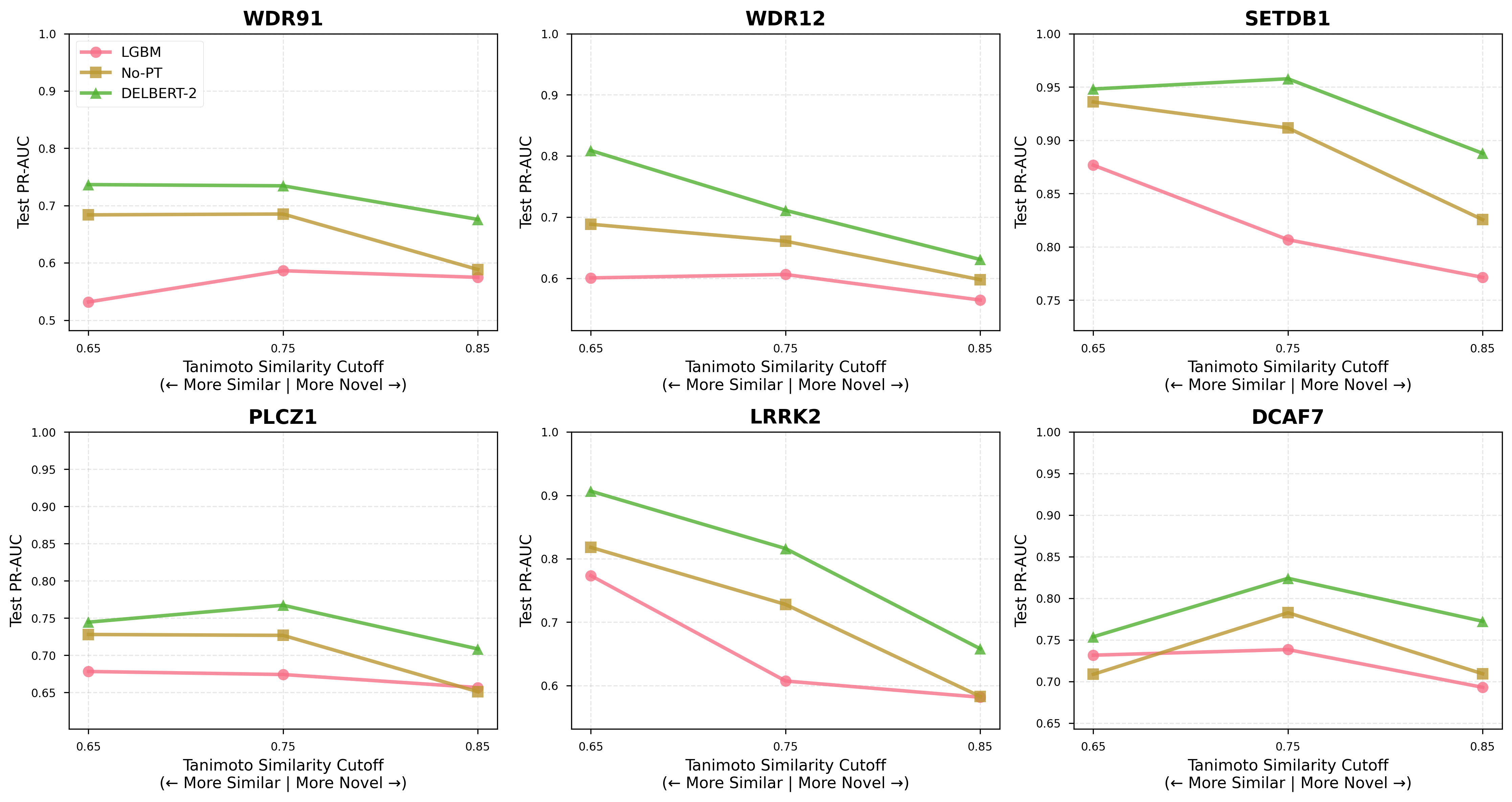}
        \caption*{(a) Per-target hierarchy split performance}
    \end{minipage}
    \hfill
    \begin{minipage}{0.40\textwidth}
        \raggedright \includegraphics[width=0.83\linewidth]{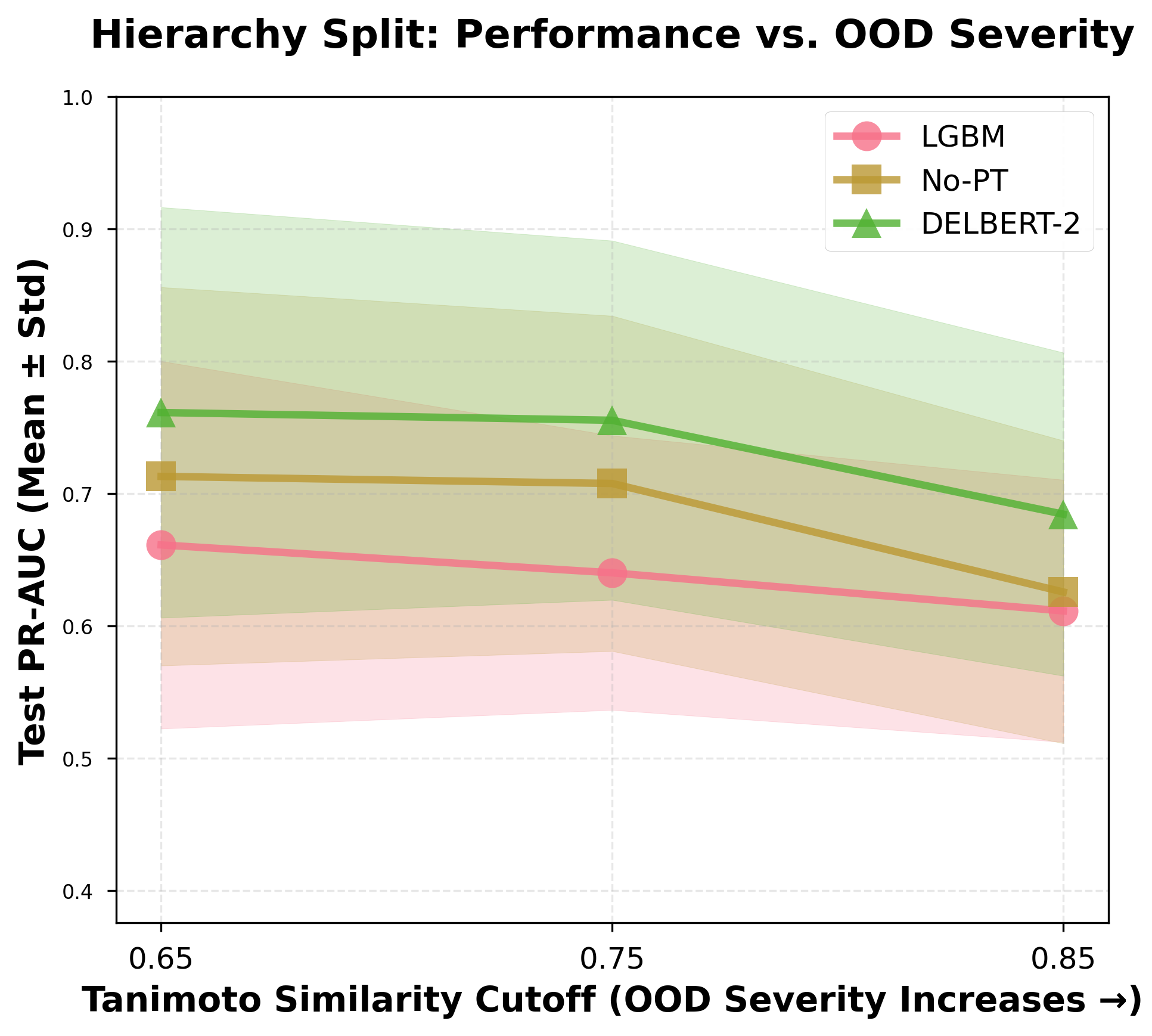}
        \caption*{(b) Mean across targets}
    \end{minipage}
    \caption{Hierarchy split performance vs.\ OOD severity (Tanimoto similarity cutoff). \textbf{(a)} Per-target Test PR-AUC degrades as the cutoff increases (0.65 $\rightarrow$ 0.85), reflecting increasing chemical novelty. DELBERT-2 consistently outperforms LGBM and No-PT across all targets. \textbf{(b)} Mean $\pm$ std across 6 targets: DELBERT-2 maintains higher performance at all severity levels, with the largest advantage at the strictest cutoff (0.85), demonstrating that unified pre-training improves robustness to chemical novelty.}
    \label{fig:hierarchy_degradation}
\end{figure*}

\begin{table*}[h]
\centering
\caption{Hierarchy split results. \textbf{Left:} PR-AUC. \textbf{Right:} NDCG@1000. Higher is better. OOD severity increases with a higher clustering cutoff. DELBERT-2 consistently outperforms LGBM and NPT across targets, with the largest gains in stricter OOD regimes.}
\label{tab:hier_pr_auc_ndcg}
\small
\setlength{\tabcolsep}{4pt}
\begin{minipage}{0.49\textwidth}
\raggedleft
\begin{tabular}{llcccccc}
\toprule
\multicolumn{2}{c}{} & \multicolumn{6}{c}{\textbf{PR-AUC}}\\
\cmidrule(lr){3-8}
\multicolumn{2}{c}{} & \multicolumn{2}{c}{Cutoff 0.65} & \multicolumn{2}{c}{Cutoff 0.75} & \multicolumn{2}{c}{Cutoff 0.85} \\
\cmidrule(lr){3-8}
Data & Model & ID & OOD & ID & OOD & ID & OOD \\
\midrule
\multirow{3}{*}{W91} & LGBM & 0.768 & 0.532 & 0.775 & 0.586 & 0.757 & 0.575 \\
 & NPT  & 0.827 & 0.684 & 0.828 & 0.685 & 0.784 & 0.588 \\
 & DBT-2 & \textbf{0.881} & \textbf{0.737} & \textbf{0.878} & \textbf{0.735} & \textbf{0.865} & \textbf{0.676} \\
\midrule
\multirow{3}{*}{W12} & LGBM & 0.689 & 0.601 & 0.713 & 0.606 & 0.681 & 0.564 \\
 & NPT  & 0.769 & 0.688 & 0.784 & 0.661 & 0.714 & 0.598 \\
 & DBT-2 & \textbf{0.812} & \textbf{0.809} & \textbf{0.821} & \textbf{0.711} & \textbf{0.749} & \textbf{0.631} \\
\midrule
\multirow{3}{*}{ST1} & LGBM & 0.898 & 0.877 & 0.896 & 0.807 & 0.902 & 0.771 \\
 & NPT  & 0.946 & 0.936 & 0.943 & 0.912 & 0.943 & 0.826 \\
 & DBT-2 & \textbf{0.976} & \textbf{0.948} & \textbf{0.973} & \textbf{0.958} & \textbf{0.962} & \textbf{0.888} \\
\midrule
\multirow{3}{*}{PL1} & LGBM & 0.719 & 0.678 & 0.703 & 0.674 & 0.673 & 0.657 \\
 & NPT  & 0.746 & 0.728 & 0.733 & 0.727 & 0.682 & 0.651 \\
 & DBT-2 & \textbf{0.786} & \textbf{0.745} & \textbf{0.767} & \textbf{0.767} & \textbf{0.709} & \textbf{0.709} \\
\midrule
\multirow{3}{*}{LK2} & LGBM & 0.719 & 0.773 & 0.718 & 0.607 & 0.691 & 0.582 \\
 & NPT  & 0.807 & 0.818 & 0.789 & 0.728 & 0.725 & 0.583 \\
 & DBT-2 & \textbf{0.879} & \textbf{0.907} & \textbf{0.866} & \textbf{0.816} & \textbf{0.806} & \textbf{0.658} \\
\midrule
\multirow{3}{*}{DC7} & LGBM & 0.823 & 0.732 & 0.824 & 0.738 & 0.828 & 0.693 \\
 & NPT  & 0.837 & 0.709 & 0.835 & 0.783 & 0.826 & 0.709 \\
 & DBT-2 & \textbf{0.851} & \textbf{0.754} & \textbf{0.852} & \textbf{0.824} & \textbf{0.852} & \textbf{0.772} \\
\bottomrule
\end{tabular}
\end{minipage}
\hfill
\begin{minipage}{0.49\textwidth}
\raggedright
\begin{tabular}{llcccccc}
\toprule
\multicolumn{2}{c}{} & \multicolumn{6}{c}{\textbf{NDCG@1000}}\\
\cmidrule(lr){3-8}
\multicolumn{2}{c}{} & \multicolumn{2}{c}{Cutoff 0.65} & \multicolumn{2}{c}{Cutoff 0.75} & \multicolumn{2}{c}{Cutoff 0.85} \\
\cmidrule(lr){3-8}
Data & Model & ID & OOD & ID & OOD & ID & OOD \\
\midrule
\multirow{3}{*}{W91} & LGBM & 0.943 & 0.832 & 0.941 & 0.779 & 0.832 & 1.000 \\
 & NPT  & 0.971 & 0.899 & 0.957 & 0.868 & 0.863 & 0.950 \\
 & DBT-2 & \textbf{0.986} & \textbf{0.916} & \textbf{0.983} & \textbf{0.883} & \textbf{0.928} & 0.997 \\
\midrule
\multirow{3}{*}{W12} & LGBM & 0.750 & 0.896 & 0.785 & 0.680 & 0.878 & 0.778 \\
 & NPT  & 0.825 & 0.916 & 0.846 & 0.727 & 0.903 & 0.828 \\
 & DBT-2 & \textbf{0.851} & \textbf{0.957} & \textbf{0.871} & \textbf{0.757} & \textbf{0.915} & \textbf{0.851} \\
\midrule
\multirow{3}{*}{ST1} & LGBM & 0.972 & 0.933 & 0.966 & 0.970 & 0.958 & 0.982 \\
 & NPT  & 0.985 & 0.971 & 0.984 & 0.987 & 0.969 & 0.935 \\
 & DBT-2 & \textbf{0.999} & \textbf{0.984} & \textbf{0.994} & \textbf{0.994} & \textbf{0.983} & \textbf{0.987} \\
\midrule
\multirow{3}{*}{PL1} & LGBM & 0.920 & 0.838 & 0.895 & 0.841 & 0.831 & 0.967 \\
 & NPT  & 0.943 & 0.865 & 0.939 & 0.935 & 0.843 & 0.854 \\
 & DBT-2 & \textbf{0.978} & \textbf{0.873} & \textbf{0.957} & \textbf{0.978} & \textbf{0.869} & \textbf{0.965} \\
\midrule
\multirow{3}{*}{LK2} & LGBM & 0.833 & 0.894 & 0.813 & 0.759 & 0.821 & 0.851 \\
 & NPT  & 0.894 & 0.915 & 0.875 & 0.860 & 0.845 & 0.774 \\
 & DBT-2 & \textbf{0.956} & \textbf{0.966} & \textbf{0.937} & \textbf{0.941} & \textbf{0.895} & \textbf{0.938} \\
\midrule
\multirow{3}{*}{DC7} & LGBM & 0.944 & 0.894 & 0.939 & 0.860 & 0.903 & 0.899 \\
 & NPT  & 0.946 & 0.901 & 0.931 & 0.900 & 0.920 & 0.868 \\
 & DBT-2 & \textbf{0.960} & \textbf{0.918} & \textbf{0.954} & \textbf{0.952} & \textbf{0.929} & \textbf{0.928} \\
\bottomrule
\end{tabular}
\end{minipage}
\end{table*}

\section{Experiments}
\label{sec:experiments}

We evaluate DELBERT-2 on DEL virtual screening with an emphasis on \emph{out-of-distribution} (OOD) generalization. 
Our evaluation probes three failure modes: compositional generalization to unseen building blocks, transfer to unseen DEL libraries, and robustness to controllable chemical novelty.
Unless stated otherwise, we compare (i) ensemble LightGBM (LGBM), (ii) the same transformer architecture without pre-training (No-PT), and (iii) DELBERT-2. Validation sets are in-distribution (ID); test sets are OOD by construction of each split protocol.

\paragraph{Datasets.}
We evaluate across six DEL fingerprint-only targets from AIRCHECK (WDR91, WDR12, SETDB1, PLCZ1, LRRK2, DCAF7). 
Each dataset contains on the order of $10^5$--$10^6$ labeled molecules with substantial class imbalance, reflecting realistic DEL hit identification conditions.
NSP was excluded from multi-panel figures due to consistently lower performance (PR-AUC $<$0.52 vs. $>$0.70 for other targets), though detailed results are included in supplementary materials.

\subsection{Validation Protocols}

Figure~\ref{fig:val_protocols} summarizes the three complementary OOD protocols used in this work.
As DEL chemistry is highly structured (e.g., through library design choices and building-block reuse), a single split strategy can substantially overestimate generalization if it preserves leakage pathways such as shared building blocks, shared libraries, or near-duplicate analogues. 
We therefore evaluate three complementary protocols that isolate different sources of distribution shift:
(i) \emph{hierarchy splits} probe robustness as chemical novelty increases;
(ii) \emph{library splits} test transfer across entire libraries with distinct synthesis designs;
and (iii) \emph{building-block splits} isolate compositional generalization when test molecules share fewer building blocks with training.
Together, these protocols approximate realistic downstream deployment, where a model must generalize to new library chemistry, new building-block combinations, and increasingly novel chemical space.

\paragraph{Metrics for DEL Virtual Screening}
\label{sec:metrics}
DEL hit prediction is both a classification task and a ranking task: in practice, a model is used to prioritize a small set of candidates for experimental follow-up.
We therefore report PR-AUC as our primary threshold-free metric, since it is sensitive to false positives under extreme class imbalance. 
To evaluate early enrichment under a realistic validation budget, we also report NDCG@1000, which discounts binders found at lower ranks and emphasizes placing true binders near the top of the ranked list.


\begin{figure*}[ht]
    \centering
    \includegraphics[width=\linewidth]{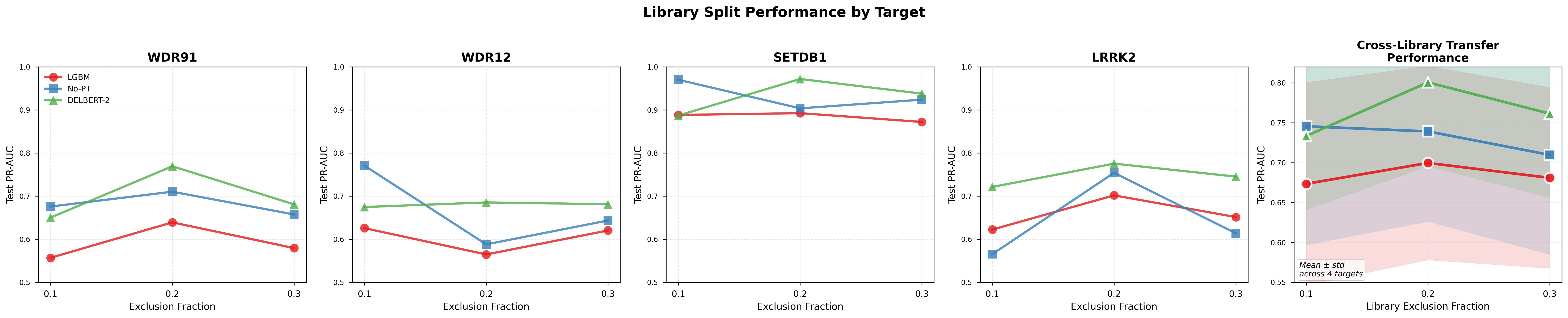}
    \caption{Cross-library transfer performance. \textbf{Left:} Per-target breakdown showing DELBERT-2 consistently above LGBM and No-PT across exclusion fractions.\textbf{Right:} Mean $\pm$ std Test PR-AUC across 4 targets (WDR91, WDR12, SETDB1, LRRK2) as the held-out library fraction increases (0.1 $\rightarrow$ 0.3). DELBERT-2 maintains more consistent performance than baselines, demonstrating better cross-library transfer from unified pre-training. }
    \label{fig:library_degradation_combined}
\end{figure*}

\begin{table*}[t]
\centering
\caption{Library split results. \textbf{Left:} PR-AUC. \textbf{Right:} NDCG@1000. Split $s\in\{0.10,0.20,0.30\}$ denotes the approximate fraction of libraries held out. DELBERT-2 consistently improves over baselines, demonstrating robust cross-library transfer.}
\label{tab:lib_pr_auc_ndcg}
\small
\setlength{\tabcolsep}{4pt}
\begin{minipage}{0.49\textwidth}
\raggedleft
\begin{tabular}{llcccccc}
\toprule
\multicolumn{2}{c}{} & \multicolumn{6}{c}{\textbf{PR-AUC}}\\
\cmidrule(lr){3-8}
\multicolumn{2}{c}{} & \multicolumn{2}{c}{Split 0.10} & \multicolumn{2}{c}{Split 0.20} & \multicolumn{2}{c}{Split 0.30} \\
\cmidrule(lr){3-8}
Data & Model & ID & OOD & ID & OOD & ID & OOD \\
\midrule
\multirow{3}{*}{LK2} & LGBM & 0.719 & 0.623 & 0.723 & 0.702 & 0.721 & 0.652 \\
 & NPT  & 0.807 & 0.565 & 0.810 & 0.754 & 0.814 & 0.614 \\
 & DBT-2 & \textbf{0.874} & \textbf{0.721} & \textbf{0.855} & \textbf{0.776} & \textbf{0.856} & \textbf{0.745} \\
\midrule
\multirow{3}{*}{ST1} & LGBM & 0.896 & 0.889 & 0.896 & 0.893 & 0.894 & 0.872 \\
 & NPT  & 0.946 & 0.970 & 0.944 & 0.904 & 0.944 & 0.924 \\
 & DBT-2 & \textbf{0.972} & 0.887 & \textbf{0.967} & \textbf{0.972} & \textbf{0.964} & \textbf{0.938} \\
\midrule
\multirow{3}{*}{W91} & LGBM & 0.779 & 0.557 & 0.773 & 0.639 & 0.769 & 0.580 \\
 & NPT  & 0.830 & 0.676 & 0.830 & 0.710 & 0.819 & 0.658 \\
 & DBT-2 & \textbf{0.885} & 0.650 & \textbf{0.884} & \textbf{0.770} & \textbf{0.887} & \textbf{0.681} \\
\midrule
\multirow{3}{*}{W12} & LGBM & 0.704 & 0.626 & 0.661 & 0.565 & 0.644 & 0.620 \\
 & NPT  & 0.772 & 0.771 & 0.762 & 0.588 & 0.732 & 0.644 \\
 & DBT-2 & \textbf{0.830} & 0.675 & \textbf{0.821} & \textbf{0.686} & \textbf{0.787} & \textbf{0.681} \\
\bottomrule
\end{tabular}
\end{minipage}
\hfill
\begin{minipage}{0.49\textwidth}
\raggedright
\begin{tabular}{llcccccc}
\toprule
\multicolumn{2}{c}{} & \multicolumn{6}{c}{\textbf{NDCG@1000}}\\
\cmidrule(lr){3-8}
\multicolumn{2}{c}{} & \multicolumn{2}{c}{Split 0.10} & \multicolumn{2}{c}{Split 0.20} & \multicolumn{2}{c}{Split 0.30} \\
\cmidrule(lr){3-8}
Data & Model & ID & OOD & ID & OOD & ID & OOD \\
\midrule
\multirow{3}{*}{LK2} & LGBM & 0.820 & 0.742 & 0.827 & 0.813 & 0.813 & 0.837 \\
 & NPT  & 0.900 & 0.716 & 0.899 & 0.815 & 0.905 & 0.773 \\
 & DBT-2 & \textbf{0.957} & 0.784 & \textbf{0.939} & \textbf{0.908} & \textbf{0.928} & \textbf{0.944} \\
\midrule
\multirow{3}{*}{ST1} & LGBM & 0.970 & 0.971 & 0.966 & 0.980 & 0.966 & 0.982 \\
 & NPT  & 0.985 & 0.988 & 0.983 & 0.987 & 0.979 & 0.980 \\
 & DBT-2 & \textbf{0.995} & \textbf{0.997} & \textbf{0.989} & \textbf{0.992} & \textbf{0.986} & \textbf{0.998} \\
\midrule
\multirow{3}{*}{W91} & LGBM & 0.946 & 0.646 & 0.936 & 0.812 & 0.915 & 0.818 \\
 & NPT  & 0.960 & 0.748 & 0.961 & 0.881 & 0.940 & 0.890 \\
 & DBT-2 & \textbf{0.978} & \textbf{0.825} & \textbf{0.979} & \textbf{0.913} & \textbf{0.979} & \textbf{0.913} \\
\midrule
\multirow{3}{*}{W12} & LGBM & 0.780 & 0.750 & 0.771 & 0.612 & 0.792 & 0.792 \\
 & NPT  & 0.857 & 0.857 & 0.861 & 0.692 & 0.872 & 0.786 \\
 & DBT-2 & \textbf{0.879} & \textbf{0.799} & \textbf{0.888} & \textbf{0.693} & \textbf{0.881} & \textbf{0.852} \\
\bottomrule
\end{tabular}
\end{minipage}
\end{table*}

\subsection{Results}
\label{sec:results}

We present results for each OOD protocol, highlighting discrimination (PR-AUC) and ranking quality (NDCG@1000). Across all protocols, DELBERT-2 generally improves over both LGBM and No-PT, with the clearest gains in stricter OOD settings.

\subsubsection{Hierarchy Cluster Split: Robustness as OOD Severity Increases}

The hierarchy split evaluates generalization under controllable distribution shift by clustering molecules in fingerprint space at different Tanimoto similarity cutoffs (0.65, 0.75, 0.85).
Higher cutoffs typically correspond to more stringent notions of chemical novelty. 
We report hierarchy split PR-AUC and NDCG@1000 in Table~\ref{tab:hier_pr_auc_ndcg}.

\paragraph{Trends with increasing OOD severity.}
Across targets, performance generally degrades as the hierarchy split becomes more stringent (i.e., as test molecules become less similar to the training distribution), which is expected for DEL virtual screening under chemical novelty.
Importantly, DELBERT-2's advantage over both LGBM and NPT is most pronounced in the stricter regimes, indicating that pretraining improves robustness rather than merely enhancing interpolation within the training distribution.


\paragraph{Practical implication.}
In real-world screening, the most valuable setting is precisely the strict OOD regime: the model must rank a small number of true binders near the top among a large candidate pool with limited overlap with the labeled DEL training data. The consistent gains in both PR-AUC and NDCG@1000 in Table~\ref{tab:hier_pr_auc_ndcg} suggest that DELBERT-2 improves not only discrimination but also early enrichment, thereby directly translating into higher expected hit yield per experimental budget.

\begin{figure*}[ht]
    \centering
    \includegraphics[width=0.95\linewidth]{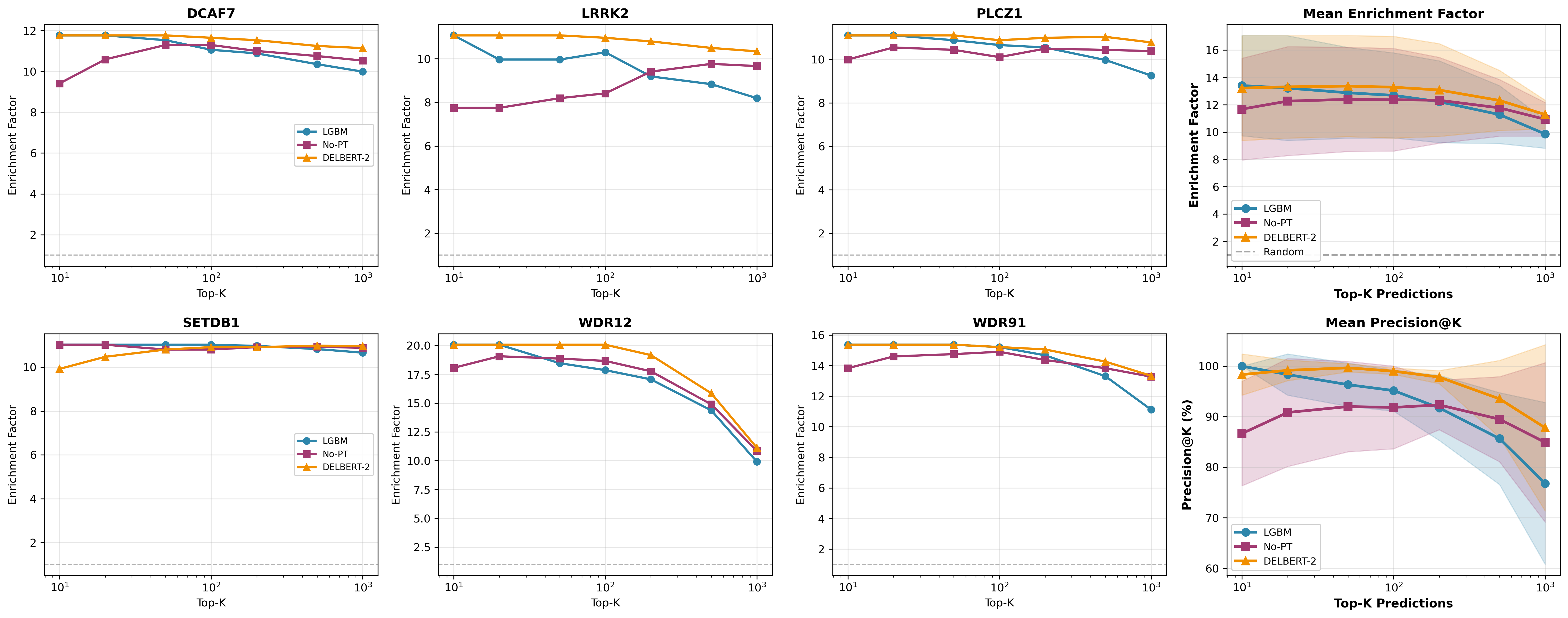}
    \caption{Enrichment analysis across targets. Top row: enrichment factors for DCAF7, LRRK2, PLCZ1, and mean across all 6 targets. Bottom row: SETDB1, WDR12, WDR91, and mean Precision@K. DELBERT-2 consistently achieves higher enrichment than baselines, with mean EF@100 of 13.28$\pm$3.72 vs.\ 12.36$\pm$3.75 (No-PT) and 12.69$\pm$3.11 (LGBM), representing 7.4\% improvement over No-PT ($p{=}0.040$). All three models are evaluated at K $\in$ \{10, 20, 50, 100, 200, 500, 1000\}.}
    \label{fig:enrichment_panels}
\end{figure*}

\begin{table*}[h]
\centering
\caption{Building-block split results. 40\% of building blocks are held out; test molecules are bucketed by overlap: \textbf{2BB} (both BBs seen), \textbf{1BB} (one seen), \textbf{0BB} (neither seen—strictest). Performance degrades as overlap decreases. DELBERT-2 often improves in stricter buckets, suggesting pre-training reduces building-block memorization. Relative: performance as \% of ID validation.}
\label{tab:bb_pr_auc_ndcg}
\small
\setlength{\tabcolsep}{3pt}
\begin{tabular}{llcccccccccccc}
\toprule
& & \multicolumn{3}{c}{\textbf{Validation}} & \multicolumn{3}{c}{\textbf{2BB Test}} & \multicolumn{3}{c}{\textbf{1BB Test}} & \multicolumn{3}{c}{\textbf{0BB Test}} \\
\cmidrule(lr){3-5} \cmidrule(lr){6-8} \cmidrule(lr){9-11} \cmidrule(lr){12-14}
\textbf{Target} & \textbf{Metric} & \textbf{LGBM} & \textbf{NPT} & \textbf{DBT-2} & \textbf{LGBM} & \textbf{NPT} & \textbf{DBT-2} & \textbf{LGBM} & \textbf{NPT} & \textbf{DBT-2} & \textbf{LGBM} & \textbf{NPT} & \textbf{DBT-2} \\
\midrule
\multirow{2}{*}{WDR91} & PR-AUC & \textbf{0.875} & 0.874 & 0.875 & 0.538 & 0.518 & \textbf{0.542} & \textbf{0.298} & 0.207 & 0.295 & \textbf{0.174} & 0.124 & 0.174 \\
 & Relative & 100\% & 100\% & 100\% & 61\% & 59\% & 62\% & 34\% & 24\% & 34\% & 20\% & 14\% & 20\% \\
\cmidrule(lr){1-14}
\multirow{2}{*}{WDR12} & PR-AUC & 0.731 & \textbf{0.798} & 0.735 & \textbf{0.481} & 0.350 & 0.461 & \textbf{0.230} & 0.159 & 0.211 & \textbf{0.161} & 0.145 & 0.142 \\
 & Relative & 100\% & 100\% & 100\% & 66\% & 44\% & 63\% & 32\% & 20\% & 29\% & 22\% & 18\% & 19\% \\
\cmidrule(lr){1-14}
\multirow{2}{*}{SETDB1} & PR-AUC & 0.943 & \textbf{0.969} & 0.932 & 0.604 & \textbf{0.655} & 0.601 & 0.293 & \textbf{0.344} & 0.309 & 0.218 & 0.187 & \textbf{0.228} \\
 & Relative & 100\% & 100\% & 100\% & 64\% & 68\% & 64\% & 31\% & 36\% & 33\% & 23\% & 19\% & \textbf{25\%} \\
\cmidrule(lr){1-14}
\multirow{2}{*}{PLCZ1} & PR-AUC & 0.772 & 0.704 & \textbf{0.776} & \textbf{0.706} & 0.629 & 0.699 & \textbf{0.624} & 0.580 & 0.611 & \textbf{0.547} & 0.540 & 0.515 \\
 & Relative & 100\% & 100\% & 100\% & 92\% & 89\% & 90\% & 81\% & 82\% & 79\% & 71\% & 77\% & 66\% \\
\cmidrule(lr){1-14}
\multirow{2}{*}{LRRK2} & PR-AUC & 0.846 & \textbf{0.878} & 0.836 & \textbf{0.636} & 0.636 & 0.610 & \textbf{0.418} & 0.351 & 0.390 & \textbf{0.223} & 0.147 & 0.220 \\
 & Relative & 100\% & 100\% & 100\% & 75\% & 72\% & 73\% & 49\% & 40\% & 47\% & 26\% & 17\% & \textbf{26\%} \\
\cmidrule(lr){1-14}
\multirow{2}{*}{DCAF7} & PR-AUC & \textbf{0.865} & 0.805 & 0.859 & 0.766 & 0.736 & \textbf{0.767} & 0.637 & 0.631 & \textbf{0.646} & 0.456 & 0.450 & \textbf{0.488} \\
 & Relative & 100\% & 100\% & 100\% & 89\% & 92\% & 89\% & 74\% & 78\% & 75\% & 53\% & 56\% & \textbf{57\%} \\
\bottomrule
\end{tabular}
\vspace{0.1cm}
\end{table*}

\begin{table*}[h]
\centering
\caption{Building-block split results. 40\% of building blocks are held out; test molecules are bucketed by overlap: \textbf{2BB} (both BBs seen), \textbf{1BB} (one seen), \textbf{0BB} (neither seen—strictest). Performance degrades as overlap decreases. DELBERT-2 often improves in stricter buckets, suggesting pre-training reduces building-block memorization. Relative: performance as \% of ID validation.}
\label{tab:bb_pr_auc_ndcg}
\small
\setlength{\tabcolsep}{3pt}
\begin{tabular}{lcccccccccccc}
\toprule
& \multicolumn{3}{c}{\textbf{Validation}} & \multicolumn{3}{c}{\textbf{2BB Test}} & \multicolumn{3}{c}{\textbf{1BB Test}} & \multicolumn{3}{c}{\textbf{0BB Test}} \\
\cmidrule(lr){2-4} \cmidrule(lr){5-7} \cmidrule(lr){8-10} \cmidrule(lr){11-13}
\textbf{Target} & \textbf{LGBM} & \textbf{NPT} & \textbf{DBT-2} & \textbf{LGBM} & \textbf{NPT} & \textbf{DBT-2} & \textbf{LGBM} & \textbf{NPT} & \textbf{DBT-2} & \textbf{LGBM} & \textbf{NPT} & \textbf{DBT-2} \\
\midrule
\multirow{1}{*}{WDR91} & \textbf{0.875} & 0.874 & 0.875 & 0.538 & 0.518 & \textbf{0.542} & \textbf{0.298} & 0.207 & 0.295 & \textbf{0.174} & 0.124 & 0.174 \\
\multirow{1}{*}{WDR12} & 0.731 & \textbf{0.798} & 0.735 & \textbf{0.481} & 0.350 & 0.461 & \textbf{0.230} & 0.159 & 0.211 & \textbf{0.161} & 0.145 & 0.142 \\
\multirow{1}{*}{SETDB1} & 0.943 & \textbf{0.969} & 0.932 & 0.604 & \textbf{0.655} & 0.601 & 0.293 & \textbf{0.344} & 0.309 & 0.218 & 0.187 & \textbf{0.228} \\
\multirow{1}{*}{PLCZ1} & 0.772 & 0.704 & \textbf{0.776} & \textbf{0.706} & 0.629 & 0.699 & \textbf{0.624} & 0.580 & 0.611 & \textbf{0.547} & 0.540 & 0.515 \\
\multirow{1}{*}{LRRK2} & 0.846 & \textbf{0.878} & 0.836 & \textbf{0.636} & 0.636 & 0.610 & \textbf{0.418} & 0.351 & 0.390 & \textbf{0.223} & 0.147 & 0.220 \\
\multirow{1}{*}{DCAF7} & \textbf{0.865} & 0.805 & 0.859 & 0.766 & 0.736 & \textbf{0.767} & 0.637 & 0.631 & \textbf{0.646} & 0.456 & 0.450 & \textbf{0.488} \\
\bottomrule
\end{tabular}
\vspace{0.1cm}
\end{table*}

\subsubsection{Library Split: Transfer to Unseen DEL Libraries}

A library split is a cross-validation protocol where we hold out \emph{entire DEL libraries} at test time. 
Each DEL dataset is a union of multiple libraries, and each library is produced by a specific combinatorial synthesis design (reaction scheme, building-block sets, and purification/selection characteristics). 
In a library split, all molecules from one or more libraries are assigned to the test fold, while training uses molecules only from the remaining libraries. 

Library splits provide a stronger and more realistic estimate of generalization than random or cluster-only splits because they reduce leakage from shared building blocks or closely related substructures across train and test sets.
By holding out whole libraries, we directly test whether a model learns transferable structure-activity relationships that generalize beyond the idiosyncrasies of particular library designs.



\paragraph{Trends with increasing OOD severity.}
We report library split PR-AUC and NDCG@1000 in Table~\ref{tab:lib_pr_auc_ndcg}.
As the held-out fraction increases (e.g., Split 0.10 $\rightarrow$ 0.30), the test set typically becomes more distribution-shifted relative to training and the task becomes harder. 
Across targets and split fractions, pretrained DELBERT-2 generally improves test PR-AUC and maintains or improves NDCG@1000, indicating better discrimination and stronger early enrichment when screening molecules drawn from unseen libraries.

\paragraph{Practical implication.}
In deployment, DEL-ML models are rarely used to re-rank close analogues from the same library chemistry used for training. 
Instead, they are used to prioritize candidates from \emph{new} libraries or external catalogues whose synthesis designs and building-block compositions differ from the labeled DEL data available for a target program. 
Library-split performance therefore better reflects real-world utility: strong results indicate that the model has learned transferable structure--activity signals that generalize beyond library-specific biases.

\subsubsection{Building-block Split: Compositional Generalization}

DEL molecules are synthesized by joining building blocks (BBs) through a fixed reaction scheme.
A building-block split is designed around how DELs are constructed.
DEL molecules are synthesized combinatorially by joining a small number of building blocks through a fixed sequence of reactions. 
We hold out 40\% of all BBs from training; test molecules are bucketed by overlap: \texttt{0bb} (no building blocks seen in training), \texttt{1bb} (one building block seen), and \texttt{2bb} (two building blocks seen). The \texttt{0bb} bucket tests compositional generalization to molecules assembled entirely from unseen building blocks.
Note that AIRCHECK does not provide SMILES for individual BBs, only for assembled molecules, precluding BB-level similarity analysis.

\paragraph{Trends with increasing OOD severity.}
We report building-block split PR-AUC and NDCG@1000 in Table~\ref{tab:bb_pr_auc_ndcg}. As building-block overlap decreases (\texttt{2bb} $\rightarrow$ \texttt{1bb} $\rightarrow$ \texttt{0bb}), performance generally drops across all models, reflecting increasing difficulty. 
Importantly, DELBERT-2 improves PR-AUC and/or NDCG@1000 in the stricter buckets relative to LGBM and NPT, suggesting that pretraining helps reduce dependence on building-block memorization.

\paragraph{Practical implication.}
The \texttt{0bb} regime most closely matches the intended deployment setting: models must prioritize candidates from chemistry not well covered by labeled training data. 
At the same time, the consistently low absolute \texttt{0bb} scores across models highlight compositional generalization as a major bottleneck, motivating future work on stronger pretraining and split-aware objectives.

\subsubsection{Enrichment Analysis: Practical Screening Impact}

To quantify the practical value of improved OOD performance for experimental validation, we analyze enrichment factors across all six targets and three OOD protocols. Enrichment measures how many more true binders are recovered in the top-K predictions compared to random selection, directly reflecting experimental screening efficiency.

\begin{figure*}[ht]
    \centering
    \begin{minipage}{0.63\textwidth}
        \centering
        \includegraphics[width=\linewidth]{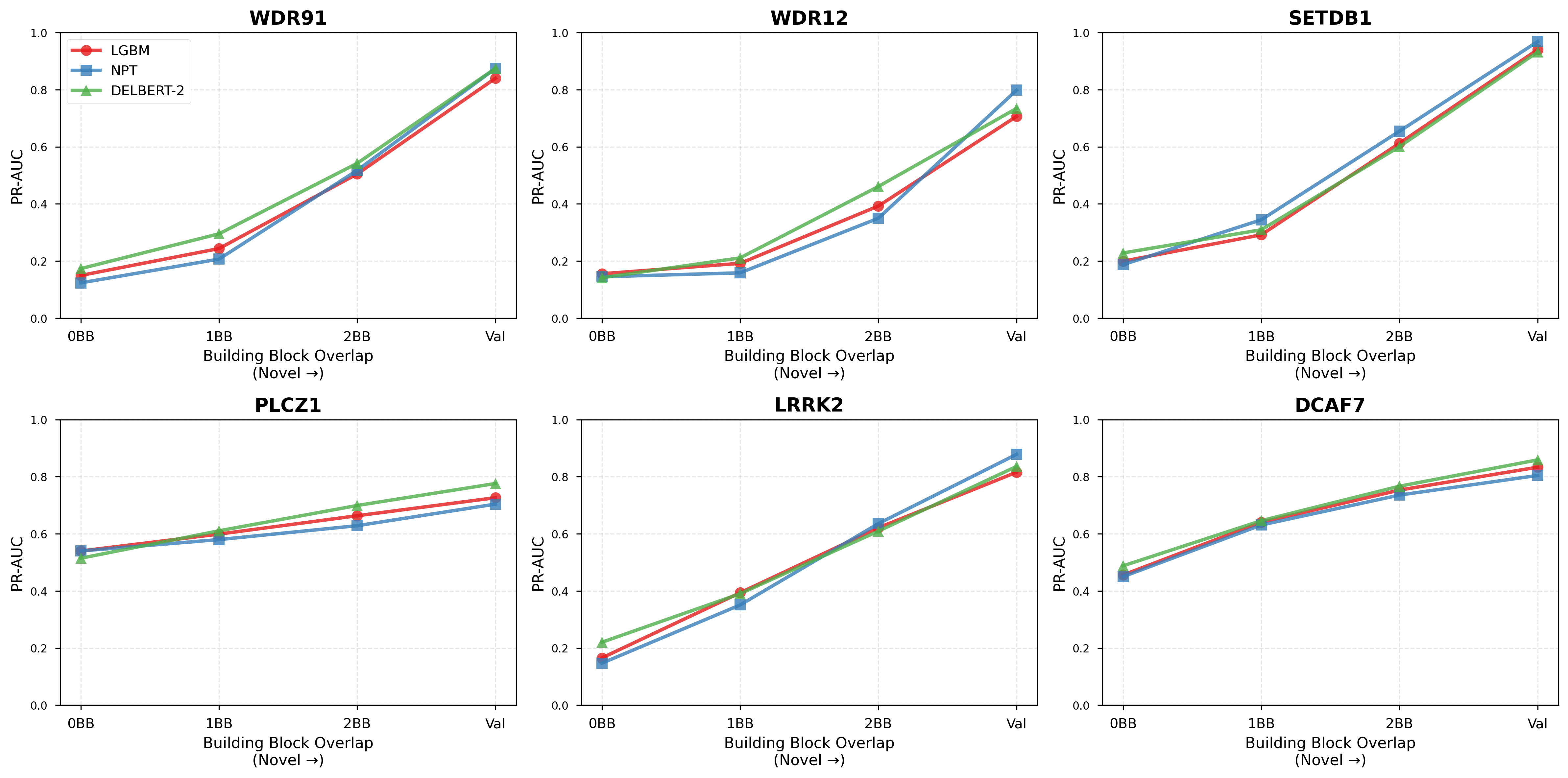}
        \caption*{(a) Mean performance across targets}
    \end{minipage}
    \hfill
    \begin{minipage}{0.33\textwidth}
        \centering
        \includegraphics[width=\linewidth]{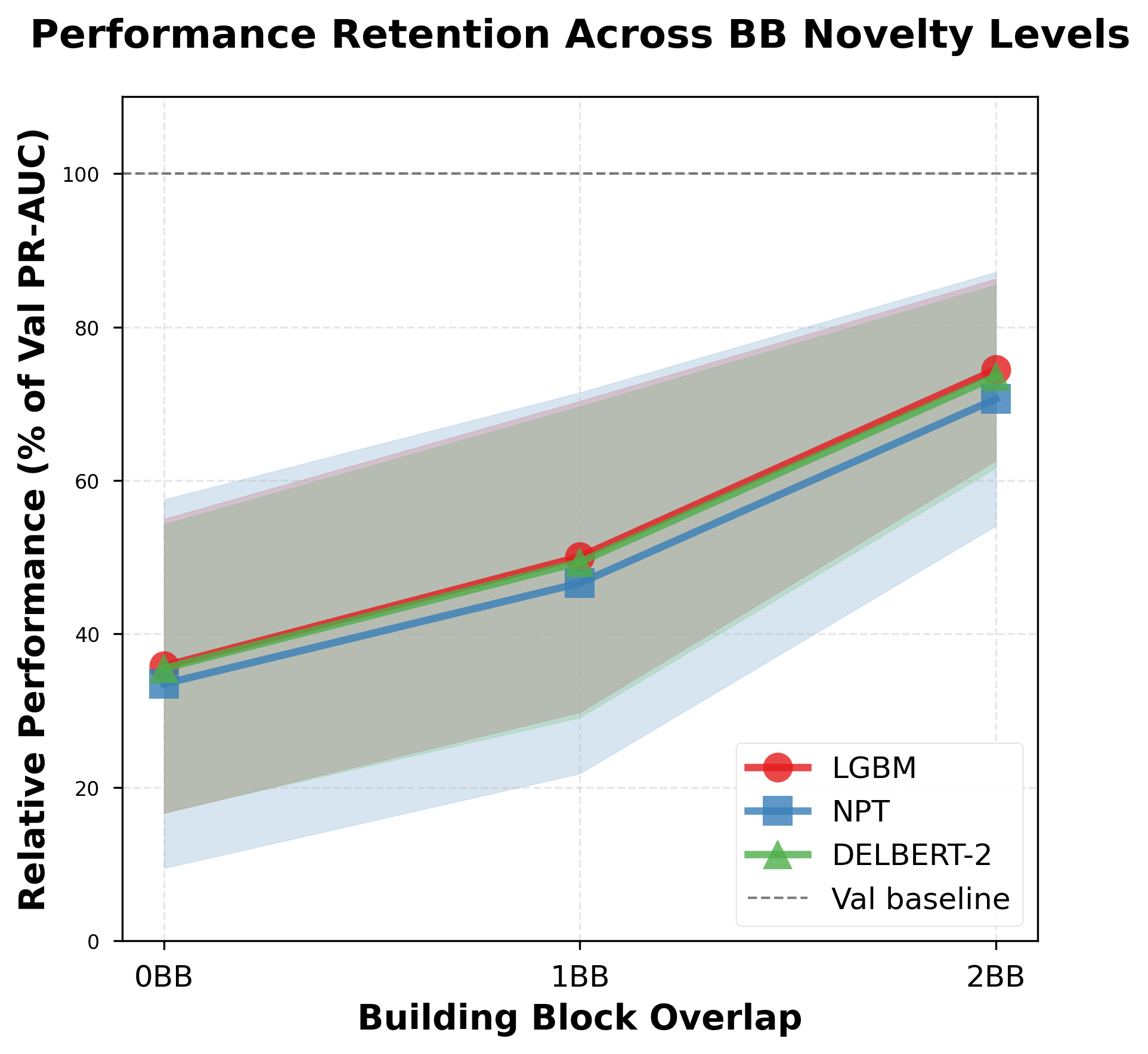}
        \caption*{(b) Performance retention}
    \end{minipage}
    \caption{Building-block compositional generalization analysis. \textbf{(a)} Test PR-AUC degrades as building-block overlap decreases (Val $\rightarrow$ 2BB $\rightarrow$ 1BB $\rightarrow$ 0BB), with DELBERT-2 showing better retention than baselines in stricter regimes. Lines show mean $\pm$ std across 6 targets. \textbf{(b)} Relative performance normalized to validation baseline. DELBERT-2 maintains higher retention in 0BB (20\%) and 1BB (34\%) buckets compared to NPT (14\% and 24\%), demonstrating that pretraining reduces building-block memorization.}
    \label{fig:bb_degradation}
\end{figure*}

Figure~\ref{fig:enrichment_panels} shows enrichment curves across representative targets and OOD protocols. DELBERT-2 consistently outperforms baselines in early enrichment (top-100 to top-500), which is critical for practical screening where experimental validation budgets are limited. Across six targets, DELBERT-2 achieves 13.28$\pm$3.72 enrichment factor at top-100 compared to 12.36$\pm$3.75 for NPT ($p{=}0.040$, paired t-test), representing a 7.4\% improvement. This translates directly to reduced screening costs: prioritizing the top 100 compounds predicted by DELBERT-2 captures approximately 13$\times$ more true binders than random selection, enabling substantial reduction in experimental validation burden while maintaining high recall.

\paragraph{Cross-protocol consistency.}
The enrichment advantage is maintained across hierarchy, library, and building-block splits, demonstrating that pretraining benefits are robust to different types of distribution shift. Even in the strictest 0BB regime where absolute performance is lowest, DELBERT-2 maintains higher enrichment than NPT, suggesting that the learned representations remain useful for hit prioritization even when compositional generalization is limited.

\section{Discussion}


A key contribution of DELBERT-2 is the unified pretraining strategy: a single model is pretrained on 2.5M molecules spanning all available AIRCHECK DEL libraries, rather than training separate per-target models.
This approach enables the model to learn generalizable molecular patterns that transfer across diverse protein targets and library designs, as evidenced by consistent improvements across six high-quality targets in Table~\ref{tab:hier_pr_auc_ndcg}.


The three validation protocols—hierarchy, library, and building-block splits—systematically probe different sources of distribution shift that arise in practical DEL virtual screening.
Hierarchy splits measure robustness to chemical novelty; library splits test cross-library transfer; building-block splits assess compositional generalization.
Together, these protocols provide a comprehensive characterization of OOD performance that cannot be captured by a single split strategy.


Across all three protocols, DELBERT-2 shows the largest gains over baselines in stricter OOD regimes (higher hierarchy thresholds, larger library exclusions, lower building-block overlap).
This suggests that self-supervised pretraining learns transferable molecular representations that reduce reliance on memorizing specific training examples, building blocks, or library-specific patterns.

\section{Limitations and Future Work}

While DELBERT-2 demonstrates strong performance, several limitations remain. 
Compositional generalization remains challenging: absolute performance in the strictest building-block regime (0BB) is modest across all models, suggesting that pretraining alone is insufficient to fully address this failure mode. 

\paragraph{Building-Block Compositional Generalization.}
Compositional generalization remains the primary limitation: absolute 0BB performance is modest across all models (PR-AUC 0.12--0.55), indicating that pre-training on assembled DEL molecules alone is insufficient to generalize to entirely unseen building blocks.
As the model has no explicit knowledge of structural or pharmacophoric relationships between building blocks, it cannot transfer what it has learned about one BB to a chemically similar but unseen BB.

\paragraph{Multi-modal Architecture.}
Future work should explore multi-modal architectures that jointly model SMILES strings and fingerprints, which may capture complementary structural information. 
By combining SMILES tokenization with fingerprint language modeling, we can leverage both the interpretability of fingerprints and the rich sequential information in SMILES.

\paragraph{Explore Auxiliary Tasks.}
An auxiliary task predicting MACCS structural keys (167 binary features) provides explicit supervision on chemically meaningful substructures, potentially improving transfer to rare scaffolds.
Preliminary experiments suggest multi-modal pretraining may reduce OOD degradation in building-block splits by learning more compositionally generalizable representations.

\paragraph{Expanded Pretraining Corpora.}
Current pretraining uses 2.5M DEL-specific molecules from AIRCHECK.
Expanding to broader chemical databases (ChEMBL: 2.4M bioactive compounds, PubChem: 10M+ structures, MOSES: 1.9M drug-like molecules) could improve coverage of molecular scaffolds underrepresented in DEL libraries.
This larger-scale pretraining (10M+ molecules) may particularly benefit generalization to novel chemical space in hierarchy and building-block splits.

Finally, prospective experimental validation in real-world screening campaigns with orthogonal binding assays and hit-to-lead optimization would establish whether computational OOD gains translate to improved experimental hit rates and reduced validation costs.

\section{Conclusion}

We introduced DELBERT-2, a pretrained fingerprint language model that converts sparse molecular fingerprints into unified token sequences and leverages ModernBERT self-supervised pretraining on 2.5M molecules from the entire AIRCHECK DEL corpus. Through unified pretraining, DELBERT-2 learns transferable molecular representations that generalize across six protein targets under three complementary out-of-distribution protocols: hierarchy splits probe chemical novelty, library splits test cross-library transfer, and building-block splits assess compositional generalization. Across these settings, DELBERT-2 consistently outperforms strong LightGBM ensemble baselines and comparable transformers trained from scratch, with particularly clear gains in stricter OOD regimes. 
These results demonstrate that fingerprint-centric self-supervised learning can effectively leverage unlabeled chemical structure data to improve hit prioritization under distribution shift, enabling practical DEL virtual screening deployment. 
More broadly, our findings highlight the need for comprehensive OOD evaluation for molecular models intended for real-world drug discovery applications.

\bibliography{example_paper}
\bibliographystyle{icml2026}

\newpage
\appendix
\onecolumn
\label{sec:appendix}

\section{Fine-tuning Hyperparameters}
\label{appendix:hp}

Table~\ref{tab:hyperparams} lists all fine-tuning hyperparameters used for DELBERT-2.
LoRA (Low-Rank Adaptation) freezes the pre-trained model weights and injects trainable low-rank decomposition matrices into the attention layers, enabling efficient fine-tuning with far fewer parameters than full fine-tuning.
Focal loss down-weights easy negatives and focuses training on hard-to-classify examples, which is critical given the 5--11\% positive class rate in DEL datasets.

\begin{table}[h]
\centering
\caption{Fine-tuning hyperparameters for DELBERT-2.}
\label{tab:hyperparams}
\begin{tabular}{ll}
\toprule
\textbf{Hyperparameter} & \textbf{Value} \\
\midrule
LoRA rank & 8 \\
LoRA $\alpha$ & 16 \\
LoRA dropout & 0.1 \\
LoRA target module & \texttt{Wqkv} (attention) \\
Focal loss $\gamma$ & 2 \\
Optimizer & AdamW \\
Early stopping metric & Validation PR-AUC \\
Positive class rate & 5--11\% (target-dependent) \\
\bottomrule
\end{tabular}
\end{table}

\section{DELBERT vs.\ DELBERT-2 Architecture Comparison}
\label{appendix:delbert_comparison}

Table~\ref{tab:model_comparison} summarizes the key architectural and training differences between the original DELBERT \cite{delbert} and DELBERT-2.
Both models use ModernBERT \cite{warner2024smarterbetterfasterlonger} as their backbone, with Flash Attention, RoPE, and alternating local-global attention. The key differences are model scale (DELBERT-2 is larger), fingerprint coverage (8 vs.\ 4 types), pre-training scope (unified 2.5M vs.\ per-target 500K), and fine-tuning strategy (focal loss + LoRA rank 8 vs.\ weighted BCE + LoRA rank 16).

\begin{table}[h]
\centering
\caption{Architectural and training comparison between DELBERT \cite{delbert} and DELBERT-2. Both use ModernBERT with Flash Attention and RoPE. Key improvements in DELBERT-2 include larger model capacity, expanded fingerprint coverage, unified pre-training on 5$\times$ more molecules, and comprehensive OOD evaluation.}
\label{tab:model_comparison}
\begin{tabular}{lll}
\toprule
\textbf{Component} & \textbf{DELBERT} & \textbf{DELBERT-2} \\
\midrule
Backbone & ModernBERT & ModernBERT \\
Parameters & $\sim$70M & $\sim$100M \\
Hidden size & 640 & 1024 \\
Layers & 16 & 12 \\
Attention heads & 16 & 16 \\
Feed-forward size & 1,152 & 1,152 \\
Attention type & Alternating local-global (Flash Attention) & Alternating local-global (Flash Attention) \\
Position encoding & Rotary (RoPE) & Rotary (RoPE) \\
Fingerprint types & 4 (ECFP4, FCFP6, AtomPair, TopTor) & 7  \\
Pre-training corpus & $\sim$500K molecules (per-target, 100 epochs) & $\sim$2.5M molecules (20 epochs) \\
Pre-training LR & $5 \times 10^{-4}$ & $1 \times 10^{-4}$ \\
Pre-training objective & MLM (15\% masking) & MLM (15\% masking) \\
LoRA rank / $\alpha$ & 16 / 32 & 8 / 16 \\
Loss function & Weighted BCE (pos\_weight=15) & Focal loss ($\gamma{=}2$) \\
OOD protocols evaluated & Library splits & Hierarchy + Library + Building-block\\
Evaluation Target Datasets & 4 targets & 6 targets\\
\bottomrule
\end{tabular}
\end{table}

\subsection*{Library Split Performance Comparison}

Table~\ref{tab:delbert_lib_comparison} compares library split PR-AUC on the four overlapping targets. For DELBERT, we report the mean $\pm$ std across all K-fold CV folds (with per-target fold counts $K$ in parentheses). For DELBERT-2, we report the OOD test PR-AUC at all three library exclusion levels (Split 0.10, 0.20, 0.30), enabling a fuller picture of how DELBERT-2 performance varies with the fraction of libraries held out.

\begin{table}[h]
\centering
\caption{Library split PR-AUC comparison between DELBERT \cite{delbert} and DELBERT-2 on the four overlapping targets. DELBERT values are mean $\pm$ std across full K-fold library CV (all libraries cycled as test); $K$ = number of CV folds per target. DELBERT-2 values are OOD test PR-AUC at three library exclusion levels (10\%, 20\%, 30\%). \textbf{Note:} Baseline models differ: DELBERT uses XGBoost+LightGBM ensembles on binarized FPs (20 models/fold); DELBERT-2 uses LightGBM ensembles on count-based FPs (5 models/fold). DCAF7 is not available in DELBERT-2's library split evaluation.}
\label{tab:delbert_lib_comparison}
\small
\setlength{\tabcolsep}{4pt}
\begin{tabular}{lccccccccc}
\toprule
& \multicolumn{3}{c}{\textbf{DELBERT OOD PR-AUC}} & \multicolumn{6}{c}{\textbf{DELBERT-2 OOD PR-AUC}} \\
\cmidrule(lr){2-4} \cmidrule(lr){5-10}
& \multicolumn{3}{c}{full K-fold CV} & \multicolumn{2}{c}{Split 0.10} & \multicolumn{2}{c}{Split 0.20} & \multicolumn{2}{c}{Split 0.30} \\
\cmidrule(lr){2-4} \cmidrule(lr){5-6} \cmidrule(lr){7-8} \cmidrule(lr){9-10}
\textbf{Target} & $K$ & \textbf{Binary Baselines} & \textbf{DBT} & \textbf{LGBM} & \textbf{DBT-2} & \textbf{LGBM} & \textbf{DBT-2} & \textbf{LGBM} & \textbf{DBT-2} \\
\midrule
WDR91  & 20 & 0.219{$\pm$0.150} & \textbf{0.349{$\pm$0.162}} & 0.557 & 0.650 & 0.639 & \textbf{0.770} & 0.580 & \textbf{0.681} \\
LRRK2  & 15 & 0.198{$\pm$0.153} & \textbf{0.336{$\pm$0.215}} & 0.623 & \textbf{0.721} & 0.702 & \textbf{0.776} & 0.652 & \textbf{0.745} \\
SETDB1 & 11 & 0.240{$\pm$0.173} & \textbf{0.585{$\pm$0.250}} & 0.889 & 0.887 & 0.893 & \textbf{0.972} & 0.872 & \textbf{0.938} \\
DCAF7  & 18 & 0.430{$\pm$0.183} & 0.405{$\pm$0.192} & - & - & -  & - & - & - \\
\bottomrule
\end{tabular}
\end{table}

\paragraph{Differences in baseline performance.}
The most striking difference between the two papers is the absolute level of baseline performance. 
This large gap is primarily attributable to the evaluation protocol: DELBERT's full K-fold CV includes many folds where the held-out library is chemically very distant from all training libraries, producing low absolute scores. 
DELBERT-2's 20\% holdout is less extreme on average. 
Additionally, DELBERT-2's baselines use count-based FPs (rather than binarized) and a larger expanded fingerprint set (8 types), which directly improves baseline performance.


\paragraph{Consistency of findings.}
Despite the protocol differences, both papers reach consistent qualitative conclusions: (i) self-supervised pretraining on fingerprint tokens substantially improves over tree-based baselines under library-based OOD evaluation; (ii) gains are largest on WDR91, LRRK2, and SETDB1, with DCAF7 being the exception where baselines are already competitive; and (iii) the improvement is most pronounced in early-enrichment metrics (EF@100, Prec@100, BEDROC in DELBERT; EF@100 in DELBERT-2), which are the most practically relevant for virtual screening.





\section{Pre-training Configuration}
\label{sec:pretrain_config}

Table~\ref{tab:pretrain_config} lists the full pre-training hyperparameters for DELBERT-2. Pre-training uses standard masked language modeling (MLM) with uniform random masking. Several optional features available in the codebase (span masking, segment-aware masking, MACCS auxiliary loss) were not used in the reported experiments.

\begin{table}[h]
\centering
\caption{Pre-training hyperparameters for DELBERT-2.}
\label{tab:pretrain_config}
\begin{tabular}{ll}
\toprule
\textbf{Hyperparameter} & \textbf{Value} \\
\midrule
MLM masking probability & 15\% \\
Batch size (per GPU) & 8 \\
Gradient accumulation steps & 8 (effective batch = 64) \\
Learning rate & $1 \times 10^{-4}$ \\
Optimizer & AdamW \\
Weight decay & 0.01 \\
Warmup ratio & 10\% of total steps \\
LR schedule & Cosine annealing (min ratio 0.1) \\
Gradient clipping & 1.0 \\
Precision & bf16-mixed \\
Pre-training epochs & 20 \\
Pre-training corpus & $\approx$2.5M molecules \\
\bottomrule
\end{tabular}
\end{table}


\section{Hierarchy Split Algorithm}
\label{sec:hierarchy_algorithm}

The hierarchical cluster split is the most technically novel evaluation protocol in this work. It proceeds in two phases: \emph{cluster construction} (run once during data preparation) and \emph{OOD sampling} (run at cross-validation time with different random seeds).

\paragraph{Cluster Construction.}
For each Tanimoto similarity cutoff $\tau \in \{0.85, 0.75, 0.65\}$, we apply the LeaderPicker algorithm \citep{rdkit} to the ECFP4 fingerprints of all molecules in the dataset:
This produces three independent cluster assignments at cutoffs 0.85 (coarse, $\sim$300 clusters), 0.75 (medium, $\sim$4K clusters), and 0.65 (fine, $\sim$30K clusters).

\paragraph{Hierarchy Rectification.}
Raw clusters at different cutoffs are not guaranteed to be nested. We apply a \emph{rectification} step: for each fine-grained cluster, we find the majority-vote coarse cluster (the coarse cluster containing the most members of the fine cluster) and reassign all members accordingly. This ensures that fine clusters are strict subsets of coarse clusters, enabling hierarchical OOD propagation. 

\paragraph{OOD Sampling.}
Given a random seed (different seeds produce different CV folds):
Start with all top-level (coarse, $\tau{=}0.85$) clusters. Sample 50\% as OOD $\rightarrow$ these molecules form the test set for cutoff 0.85.
\textbf{Propagate}: collect only the medium-level ($\tau{=}0.75$) child clusters that belong to the OOD coarse clusters. Sample 25\% of these as OOD $\rightarrow$ test set for cutoff 0.75.
\textbf{Propagate again}: collect only the fine-level ($\tau{=}0.65$) child clusters of the OOD medium clusters. Sample 10\% as OOD $\rightarrow$ test set for cutoff 0.65.
Non-OOD molecules are split 90/10 into train/val using stratified sampling.

\paragraph{Nesting Property.}
By construction, the OOD test sets are \emph{nested}: the $\tau{=}0.85$ test set $\subseteq$ the $\tau{=}0.75$ test set $\subseteq$ the $\tau{=}0.65$ test set. This enables controlled comparison across OOD severity levels without confounding by test set composition — the same molecules appear in all three test sets, with additional molecules added at looser cutoffs.




\section{Building-block Split Details}
\label{sec:bb_split_details}

The building-block split uses a \emph{label-aware} holdout strategy to ensure the held-out building blocks include BBs that appear in positive (binding) molecules, preventing trivially easy splits where all positive molecules share training BBs.

\paragraph{Label-aware BB sampling.} Of the 40\% of BBs held out: 50\% are sampled from BBs that appear in at least one positive molecule (binder) 50\% are sampled randomly from the remaining BBs.
This ensures the held-out set contains pharmacophoric BBs, making the 0BB test set genuinely challenging.
\paragraph{Train/test assignment.}
Each molecule is assigned a BB overlap score $k \in \{0, 1, 2, 3\}$ = number of its building blocks present in the training BB set: $k{=}3$: all BBs in training is validation; $k{=}2$: 2 BBs shared is medium OOD; $k{=}1$: 1 BB shared is high OOD; $k{=}0$: no BBs shared is strictest OOD.

\paragraph{Dataset limitation.}
AIRCHECK does not provide SMILES or structural information for individual building blocks; only for the assembled DEL molecules. This precludes BB-level similarity analysis (e.g., testing whether models generalize better to structurally similar held-out BBs). Future work with datasets that expose BB structures could enable finer-grained compositional generalization studies.


\section{Library Split Details}
\label{sec:library_split_details}

\paragraph{Library prefix extraction.}
Each molecule's \texttt{LIBRARY\_ID} encodes its library of origin (e.g., \texttt{L34\_001} $\rightarrow$ library \texttt{L34}). We extract the library prefix by stripping the last underscore-delimited component, grouping all molecules from the same synthesis design.\textbf{Small library handling.}
Libraries with fewer than 10 molecules are merged into a combined group that always remains in training. This prevents degenerate test sets from very small libraries while preserving the full dataset for training.
\textbf{Exclusion percentages.}
We evaluate at three exclusion fractions: 10\%, 20\%, and 30\% of large libraries held out. Each exclusion level uses a deterministic seed derived from the base CV seed, ensuring reproducibility. No library appears in both train and test sets.


\end{document}